\documentclass[runningheads]{llncs}
\usepackage{graphicx}
\usepackage{xcolor,colortbl}
\usepackage{multirow,adjustbox}
\usepackage{subfigure}
\definecolor{LightCyan}{rgb}{0.88,1,1}
\definecolor{LightGreen}{rgb}{0.8,1,0.6}
\definecolor{LightBlue}{rgb}{0.5,0.9,1}
\definecolor{LightPurple}{rgb}{0.9,0.8,1}
\makeatletter
\newcommand{\printfnsymbol}[1]{%
  \textsuperscript{\@fnsymbol{#1}}%
}
\makeatother

\begin{document}
\title{Predicting Drug-Drug Interactions from Heterogeneous Data: \\ An Embedding Approach}

%
\titlerunning{Predicting Drug-Drug Interactions from Heterogeneous Data}
%
\author{Devendra Singh Dhami \thanks{equal contribution} \inst{1} \and
Siwen Yan \printfnsymbol{1} \inst{2} \and
Gautam Kunapuli\inst{3} \and
David Page \inst{4} \and
Sriraam Natarajan \inst{2} }
\authorrunning{Dhami et al.}
%
\institute{Technical University of Darmstadt, Germany \and
The University of Texas at Dallas, USA \and
Verisk Analytics, USA \and
Duke Univeristy, USA}
\maketitle              
\begin{abstract}
Predicting and discovering drug-drug interactions (DDIs) using machine learning has been studied extensively. However, most of the approaches have focused on text data or textual representation of the drug structures. We present the first work that uses multiple data sources such as drug structure images, drug structure string representation and relational representation of drug relationships as the input. To this effect, we exploit the recent advances in deep networks to integrate these varied sources of inputs in predicting DDIs. Our empirical evaluation against several state-of-the-art methods using standalone different data types for drugs clearly demonstrate the efficacy of combining heterogeneous data in predicting DDIs.

\end{abstract}

\section{Introduction}
Adverse drug events (ADEs) are ``injuries resulting from medical intervention related to a drug''~\cite{NebekerBS04}, and are distinct from medication errors (inappropriate prescription, dispensing, usage etc.). 
ADEs can account for as many as one-third of hospital-related complications, affect up to 2 million hospital stays annually, and prolong hospital stays by 2--5 days \cite{HHS10}.
Recently it is observed that many of these ADEs can be attributed to {\bf very common medications} \cite{BudnitzEtAl11} and {\bf many are preventable} \cite{GurwitzEtAl03} or {\bf ameliorable} \cite{ForsterEtAl05}. 

We focus on a specific problem of {\bf drug-drug interactions} (DDIs), which are an important type of ADE and can potentially result in healthcare overload or even death~ \cite{becker2007hospitalisations}. An ADE is characterized as a DDI when multiple medications are co-administered and cause an adverse effect on the patient. 
Predicting and discovering drug-drug interactions (DDIs) is an important problem and has been studied extensively both from medical and machine learning point of view. Identifying DDIs is an important task during drug design and testing, and several regulatory agencies require large controlled clinical trials before approval. Beyond their expense and time-consuming nature, it is impossible to discover all possible interactions during such clinical trials. This necessitates the need for computational methods for DDI prediction. A substantial amount of work in DDI focuses on homogeneous data types such as text \cite{liu2013azdrugminer,chee2011predicting}, textual representation of the structural data of drugs \cite{gurulingappa2012extraction,asada2018enhancing} and genetic data \cite{qian2019leveraging}. Recent approaches consider phenotypic, therapeutic, structural, genomic and reactive drug properties \cite{cheng2014machine} or their combinations \cite{dhami2018drug} to characterize drug interactivity but this type of information only serves to extract {\em in vivo/vitro} discoveries. 

Our goal is to predict DDIs in large drug databases by exploiting heterogeneous data types of the drugs and identifying patterns in drug interaction behaviors. We take a {\em fresh and novel perspective on DDI prediction by seamlessly combining heterogeneous data representations} of the drug structures such as images, string representations and relations with other proteins. 
While in principle, multi-view learning methods such as co-training~\cite{blum1998combining} or multiple kernel learning~\cite{gonen2011multiple} can be used, these methods assume that each view independently provides enough information for classification while we assume that each of these data source essentially provides a weak prediction of DDI. While it is possible to directly combine the data sources, standardization can be a major bottleneck. We take an embedding based approach to achieve the combination.

We make the following contributions: (1) we combine heterogeneous data types representing drug structures for DDI prediction. (2) we create embeddings to build a DDI prediction engine that can be integrated into a drug database seamlessly. (3) we show that using heterogeneous data types is more informative than using homogeneous data types.

\section{Related Work}
While DDIs have been long explored from medical perspective~ \cite{lau2003atorvastatin,hirano2006drug,becker2007hospitalisations,bjorkman2002drug}, or from social and economic perspectives~\cite{arnold2018impact,shad2001economic}, we take a machine learning approach to this task. 

Classically, the task of DDI discovery/prediction is modeled as a pairwise classification task. Thus kernel-based methods \cite{Shawe-TaylorCristianini04} are a natural fit since kernels are naturally suited to representing pairwise similarities. Most similarity-based methods for DDI discovery/prediction construct NLP-based kernels from literature data \cite{segura2011using,chowdhury2013fbk}. A different direction is to learn kernels from different types of data such as molecular and structural properties of the drugs and then using these multiple kernels to predict DDIs \cite{cheng2014machine,dhami2018drug}.
Recently embeddings have been employed for learning from a single data source \cite{purkayastha2019drug,celebi2019evaluation}. Siamese networks have been applied in one shot image recognition ~\cite{koch2015siamese}, signature verification ~\cite{bromley1994signature}, medical question retrieval \cite{wang2019medical} and Alzheimer disease diagnosis \cite{aderghal2017classification}. For DDI, Siamese graph convolutional networks have been developed~\cite{chen2019drug,jeon2019resimnet}. Most of these work for the DDI classification considered homogeneous data source. Even when heterogeneous data sources are considered, the methods tend to qualify for finding similarity scores between various drugs and then thresholding the obtained scores for prediction. An important limitation is the exclusion of drug structure images as a type of data. Our work can be seen as {\em the first generalization of these multiple methods} where we consider multiple data sources including images and combine them seamlessly through embeddings. 

\section{Embeddings using Heterogeneous Data Sources}
\begin{figure*}[!t]
    \centering
    \includegraphics[width=\textwidth]{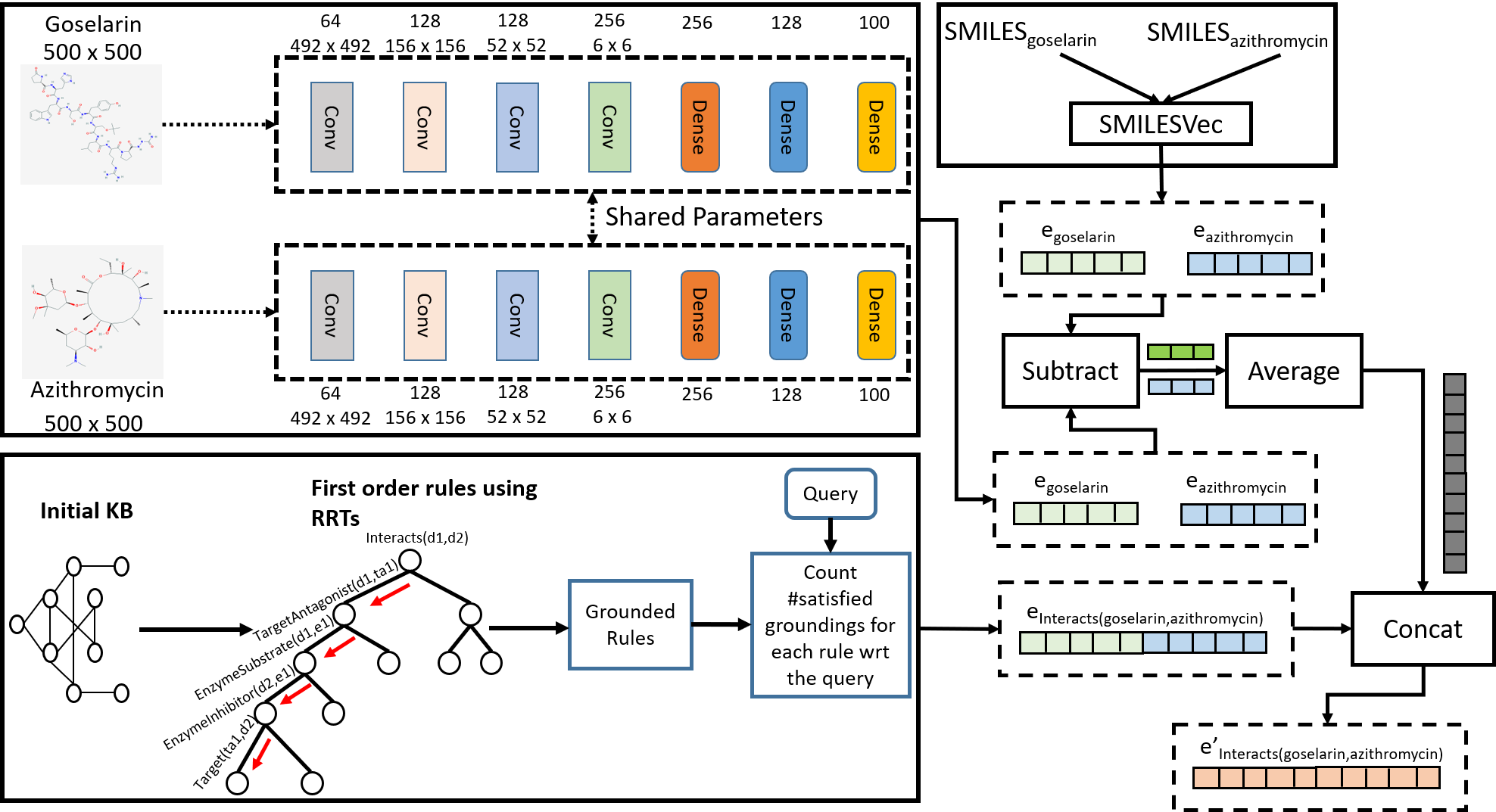}
    \caption{Overview of architecture for predicting DDIs using heterogeneous data types.}
    \label{fig:overview}
\vspace{-0.2in}
\end{figure*}

We consider 3 different types of data, (1) images of drug structures, (2) SMILES (\textbf{S}implified \textbf{M}olecular \textbf{I}nput \textbf{L}ine \textbf{E}ntry \textbf{S}ystem) strings \cite{weininger1988smiles} representation of drug structures and (3) relational representation of various associations between the drugs and proteins (target, transporter and enzymes). Figure \ref{fig:overview} shows the overall architecture of our approach. We now discuss the different components.

\vspace{-0.5em}
\subsection{Drug Structure Image Embeddings}
A discriminative approach for learning a similarity metric using a Siamese architecture ~\cite{chopra2005learning} maps the input (pair of images in our case) into a target space. The intuition is that the distance between the mappings is minimized in the target space for similar pairs of examples and maximized in case of dissimilar examples. We adapt the Siamese architecture for the task of generating embeddings for each drug image. It consists of two identical sub-networks i.e. networks having same configuration with the same parameters and weights. Each sub-network takes a gray-scale image of size $500 \times 500 \times 1$ as input (we convert colored images to gray-scale) and consists of $4$ convolutional layers with number of filters as $64$, $128$, $128$ and $256$ respectively. The kernel size for each convolutional layer is (9 $\times$ 9) and the activation function is \textit{relu}. The \textit{relu} is a non-linear activation function is given as $f(x)=max(0,x)$. Each convolutional layer is followed by a max-pooling layer with pool size of ($3 \times 3$) and a batch normalization layer. After the convolutional layers, the sub-network has $3$ fully connected layers with $256$, $128$ and $100$ neurons respectively. Each drug pair is used to train the Siamese network and the learned parameters are used to generate embeddings of dimension $100 \times 1$ for each drug image. 

\begin{figure*}[h!]
    \begin{center}
    \includegraphics[width=\textwidth]{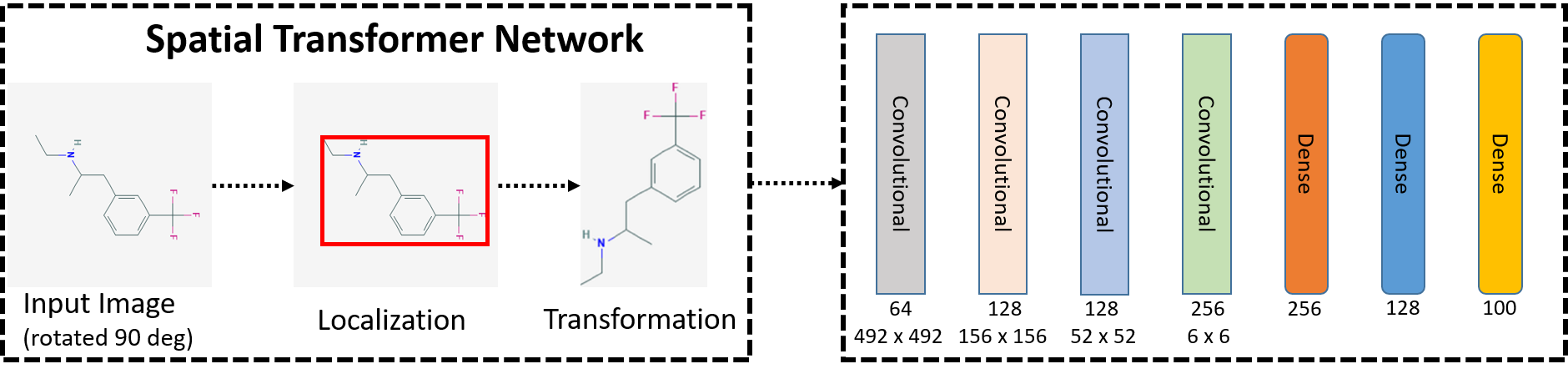}
    \end{center}
    \caption{Using spatial transformer network as a pre-processing step to mitigate rotational variance. Note that this process is done for both the input images.}
    \label{fig:stn}
    \vspace{-0.4in}
\end{figure*}

Note that the convolutions in the convolutional sub-network provide translational in-variance. However, rotational in-variance is also crucial, since isomers (one of the chiral forms) of drugs are expected to react differently when interacting with a certain drug \cite{nguyen2006chiral}. For example, Fenfluramine and Dexfenfluramine are isomers of each other but Fenfluramine interacts with Acebutolol while Dexfenfluramine does not. Thus, to introduce rotational invariance, we use spatial transformer networks (STN) \cite{jaderberg2015spatial} consisting of three basic building blocks: a localisation network, a grid generator and a sampler which can be used as a pre-processing step before feeding the input image pair into our underlying architecture (Figure \ref{fig:stn}).

\subsection{Relational Data Embeddings}

\begin{figure*}[h!]
    \begin{center}
    \includegraphics[width=\textwidth]{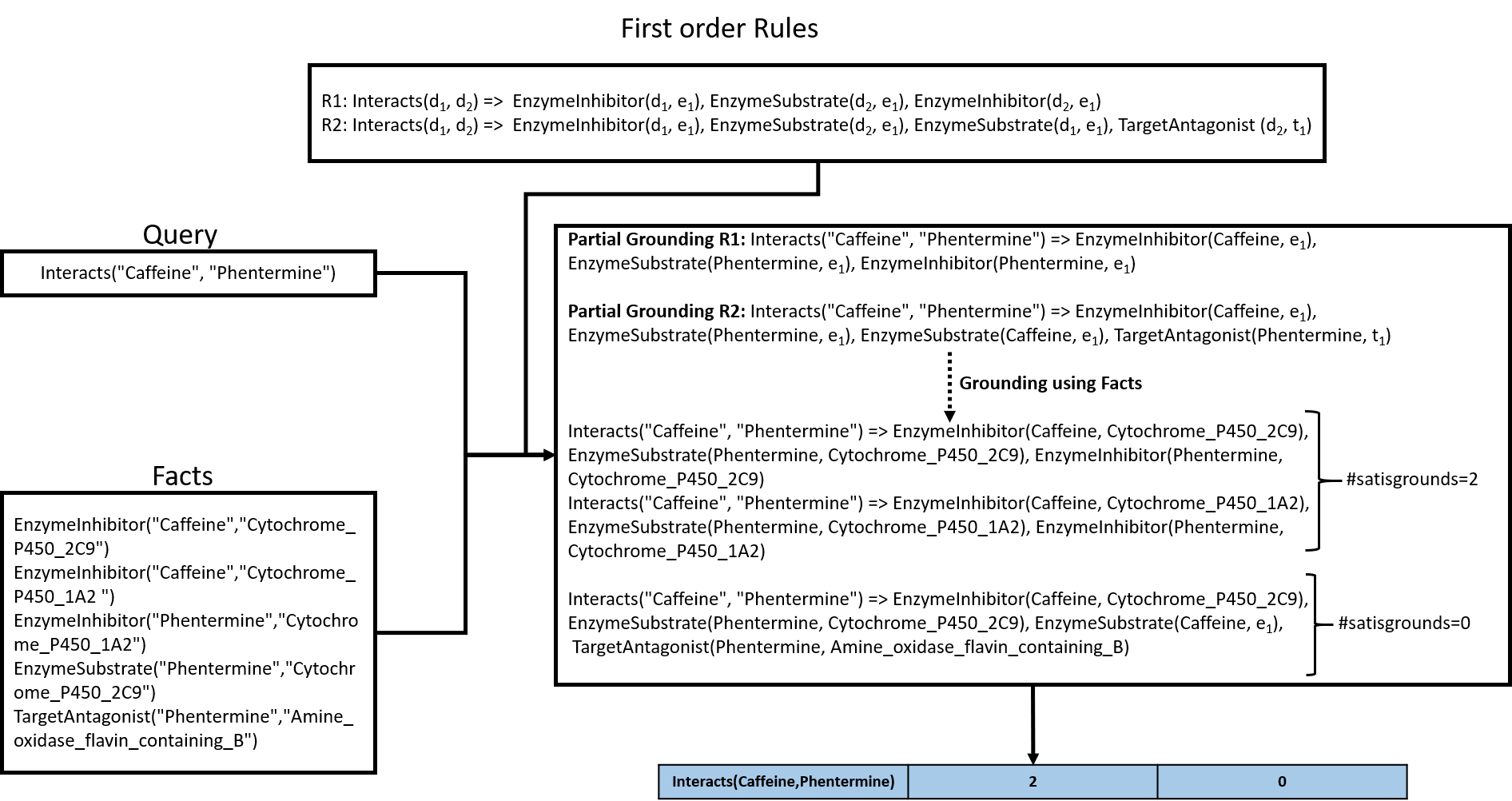}
    \end{center}
    \caption{Embedding creation from relational data.}
    \label{fig:grounding}
    \vspace{-0.2in}
\end{figure*}

DDIs can be considered as the characterization of the relationships between the drugs and the various proteins (enzymes, transporters etc.) using ADMET (absorption, distribution, metabolism, excretion and toxicity) features. A natural representation for such data is using first-order logic and the rules can then be induced. Some example facts (features) in our knowledgebase are: 
\begin{enumerate}
    \item $TargetAgonist(``Goserelin",``Gonadotropin~releasing~hormone~receptor").$
    \item $EnzymeInhibitor(``Azithromycin",``Cytochrome~P450~2A6").$
    \item $TransporterInducer(``Alfentanil",``Multidrug~resistance~protein~1").$
\end{enumerate}
Using the given facts and the +ve and -ve examples, we learn a relational regression tree (RRT) \cite{blockeel1998top} where all the paths from the root to the leaves can be interpreted as first-order rules. The obtained first-order rules are first partially grounded with the query drug pairs and then completely grounded using the fact set. The number of satisfied groundings for each drug pair are then counted to obtain the final embeddings as shown in Figure \ref{fig:grounding}. 

\subsection{Drug Structure SMILES Strings Embeddings}
SMILES strings represent the drug structure in form of a simple textual representation. For example, Fluvoxamine can be represented as \textit{COCCCCC(=NOCCN)\\C1=CC=C(C=C1)C(F)(F)F}. We use the existing model of SMILESVec \cite{Ozturk2018Anovel} which divides the SMILES string into several interacting sub-structures and then uses the word2vec method \cite{mikolov2013efficient} to generate embeddings for these sub-structures. These embeddings are combined to generate the final embedding of the drugs.

\subsection{Combining Embeddings of Heterogeneous Data}
After the embeddings of all the 3 hetrerogeneous data are obtained as described above, these need to be aggregated in order to generate a lower level representation. In the case of both image and SMILES strings embeddings (both of size $100\times1$), we hypothesize that more similar the structure of the drugs, higher is the probability of their interaction. To capture this similarity notion between both sets of embeddings, we use \textbf{subtraction} as the aggregation function to obtain 2 sets of embeddings for the image and SMILES strings data. These 2 sets are then averaged to obtain a single set of embeddings of size $100\times1$.

Each relational embedding represents the counts of the satisfied groundings of the query, in our case, $Interacts(d_1,d_2)$ i.e. the interaction between pair of drugs and is of the size $19\times1$ ($19$ is the number of first-order rules learned using the relational regression trees). The relational embeddings are concatenated with the combined embeddings obtained from the SMILES and image data to yield the final embedding size of $119\times1$ which can then be passes to a machine learning classifier. We choose a neural network since it is a universal approximator, can handle large number of features and also learns inherently aggregated latent features in the hidden layers. The over all architecure is presented in Figure~\ref{fig:overview}.

\section{Empirical Evaluation}
We aim to answer the following questions: \textbf{Q1.} Does using multiple data sources give an advantage over using a single data source? \textbf{Q2.} Does using STN in the Siamese neural network give better results? \textbf{Q3.} What is the effect of the aggregation functions? \textbf{Q4.} Is the classification performance sensitive to the choice of classifier? \textbf{Q5.} Does the size of hidden layers and different activation function in the neural classifier affect the performance? 

\textbf{Data set(s):} Our image data set consists of images of $373$ drugs of size $500$ $\times 500 \times 3$ downloaded from the PubChem database~\footnote{https://pubchem.ncbi.nlm.nih.gov/} and converted to a grayscale format of size $500\times500\times1$. The images are then normalized by the maximum pixel value (i.e. $255$). The SMILES strings of these drugs are obtained from PubChem and DrugBank~\footnote{https://www.drugbank.ca/}. For the relational data, we extract the different relations of the drugs with the proteins from DrugBank and convert it to a relational format with number of relations $= 14$ and the total number of facts $= 5366$. From the $373$ drugs we create a total of $67,360$ drug interaction pairs excluding the reciprocal pairs (i.e. if drug $d_1$ interacts with drug $d_2$ then $d_2$ interacts with $d_1$ and are removed). From the $67,360$ drug pairs we obtain $19936$ drug pairs that interact and $47424$ drug pairs that do not.

\textbf{Baselines:} We consider $7$ baselines based on the different modalities to compare the results from our architecture. \textbf{Structural Similarity Index (SSIM)} \cite{wang2004image} is used for measuring perceptual similarity between images and is calculated as, $SSIM(X_1,X_2)=\frac{(2 \mu_{X_1}  \mu_{X_2} + C_1) \times (2 \sigma_{X_1 X_2} + C_2)}{(\mu_{X_1}^2 + \mu_{X_2}^2 +C_1) \times (\sigma_{X_1}^2 + \sigma_{X_2}^2 +C_2)}$, where $\mu_{X_1}$ and $\mu_{X_2}$ is the average of the images $X_1$ and $X_2$ respectively, $\sigma_{X_1}$ and $\sigma_{X_2}$ is the variance of the images $X_1$ and $X_2$ respectively, $\sigma_{X_1 X_2}$ is the covariance of the two input images. The constants $C_1$ and $C_2$ are added to the SSIM to avoid instability. To obtain the predictions, the threshold is set as the mean SSIM values of all pairs. \textbf{Autoencoders} \cite{kramer1991nonlinear} are neural networks with an encoder that extracts features from the input images and a decoder that restores the original images from the extracted features. The autoencoder is trained for $10$ epochs with binary cross-entropy loss. The encoder extracts features of the testing images. To find images with similar extracted features $2$ criteria were used: {\em binary cross-entropy} and {\em cosine proximity}. The threshold to decide similarity of $2$ images is the mean of all values calculated for all pairs of testing image. \textbf{CASTER} \cite{huang2020caster} identifies the frequent substrings present in the SMILES strings using a sequential pattern mining algorithm which are then converted to an embedded representation using an encoder module to obtain a set of latent features which are then converted into linear coefficients, passed through a decoder and a predictor to obtain the DDI predictions. \textbf{Siamese Neural Network with and without STNs} using contrastive loss \cite{hadsell2006dimensionality}, based on a euclidean distance are also used as baselines. If  the distance between images $\geq$ 0.65 (obtained using AUC-PR curves) we predict an interaction. \textbf{RDN-Boost} \cite{natarajan2012gradient} takes an initial model (RRT) and use the obtained predictions to compute gradient(s)/residues. A new regression function is then learnt to fit the residues and the model is updated. At the end, a combination (the sum) of all the obtained regression function gives the final model. \textbf{MLN-Boost} \cite{khot2011learning} boosts the undirected Markov logic networks (MLNs) \cite{richardson2006markov} using an approximation of likelihood. 

\textbf{Results:} We optimize the Siamese network using the Adam optimizer with a learning rate of $5 \times 10^{-5}$, obtained using line search. We use the publicly available implementation\footnote{https://github.com/hkmztrk/SMILESVecProteinRepresentation} of SMILESVec method with default parameters. To learn the RRT, we use the publicly available software, BoostSRL\footnote{https://starling.utdallas.edu/software/boostsrl/}, with the ``-noBoost" parameter. For the classifier in our architecture we use a 4 hidden layer(s) neural network with hidden layer sizes 1000, 500, 200 and 50 with \textit{relu} activation units and \textit{Adam} optimizer. Table \ref{tab:results} shows the performance of our method with respect to various baselines.

\begin{table*}[!ht]
    \centering
    \caption{Comparison of our method with baselines. The 1st 4 methods use images as input, CASTER uses SMILES strings and the next 2 use relational data.}
    \label{tab:results}
    \begin{tabular}{|c|c|c|c|c|}
        \hline
         \textbf{Methods} & \textbf{Accuracy} & \textbf{Recall} & \textbf{Precision} & \textbf{F1 score}\\
        \hline
        \rowcolor{LightCyan}
        SSIM & 0.519 & 0.487 & 0.304 & 0.374\\
        \hline
        \rowcolor{LightCyan}
        Autoencoder & 0.354 & \textbf{0.911} & 0.303 & 0.454\\
        \hline
        \rowcolor{LightCyan}
        Siamese Network & 0.837 & 0.780 & 0.705 & 0.741\\
        \hline
        \rowcolor{LightCyan}
        Siamese Network + STN & 0.823 & 0.825 & 0.661 & 0.734\\
        \hline
        \rowcolor{LightPurple}
        CASTER & 0.821 & 0.663 & 0.736 & 0.698\\
        \hline
        \rowcolor{LightGreen}
        RDN-BOOST & 0.773 & 0.832 & 0.413 & 0.552\\
        \hline
        \rowcolor{LightGreen}
        MLN-BOOST & 0.767 & 0.653 & 0.540 & 0.592 \\
        \hline
        \rowcolor{LightBlue}
        \textbf{Our Method (agg=avg)} & 0.877 & 0.769 & 0.805 & 0.787\\
        \hline
        \rowcolor{LightBlue}
        \textbf{Our Method (agg=sub)} & \textbf{0.884} & 0.781 & \textbf{0.818} & \textbf{0.799}\\
        \hline
        \rowcolor{LightBlue}
        \textbf{Our Method (with STN)} & 0.881 & 0.779 & 0.811 & 0.794\\
        \hline
    \end{tabular}
\vspace{-0.2in}
\end{table*}

\textbf{(Q1) Advantage of Heterogeneous data.} To demonstrate the effectiveness of using heterogeneous data, we compare with methods that use homogeneous data. To that effect, the first 4 baselines consider the image data, CASTER uses the SMILES strings data and RDN-Boost and MLN-Boost use the relational data. The results show that combining embeddings from heterogeneous data sources clearly outperform the methods using a single data source thus answering \textbf{Q1 affirmatively} .

\textbf{(Q2) Effect of STN.} Table \ref{tab:results} also shows the result for using STN while generating the image embeddings before aggregation. The results do not show much deviation from not using STNs. Thus we can answer \textbf{Q2}. Using STN as pre-processing to generate image embeddings does not have any significant effect.

\textbf{(Q3) Effect of Aggregation Functions.} As shown in table \ref{tab:results} the subtraction aggregation for calculating set of image and SMILES embeddings performs better than the average aggregation function since, as mentioned before, these embeddings represent the similarity information and can thus be captured by subtraction aggregation although the difference is not much thus answering \textbf{Q3}.

\begin{figure*}
\centering
    \begin{minipage}{0.32\textwidth}
        \includegraphics[width = \columnwidth]{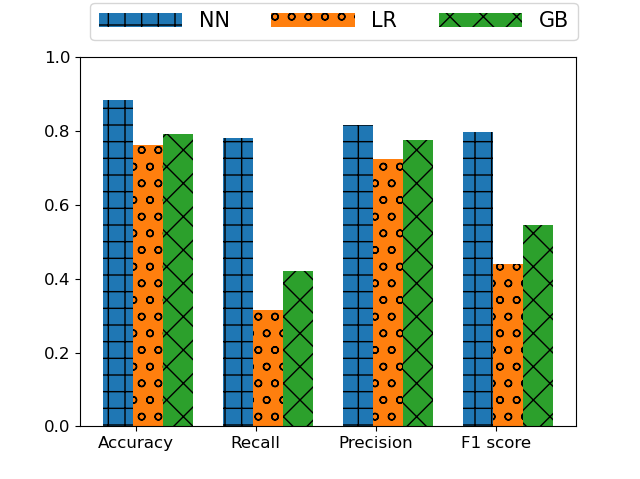}
    \caption{Effect of classifier choice}
    \label{fig:algo}
    \end{minipage}
    \centering
    \begin{minipage}{0.32\textwidth}
        \includegraphics[width = \columnwidth]{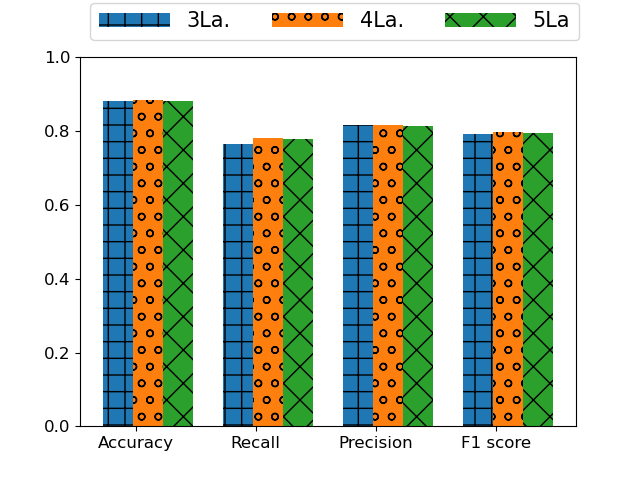}
    \caption{Effect of number of layers}
    \label{fig:layers}
    \end{minipage}
    \begin{minipage}{0.32\textwidth}
    \centering
       \includegraphics[width=\columnwidth]{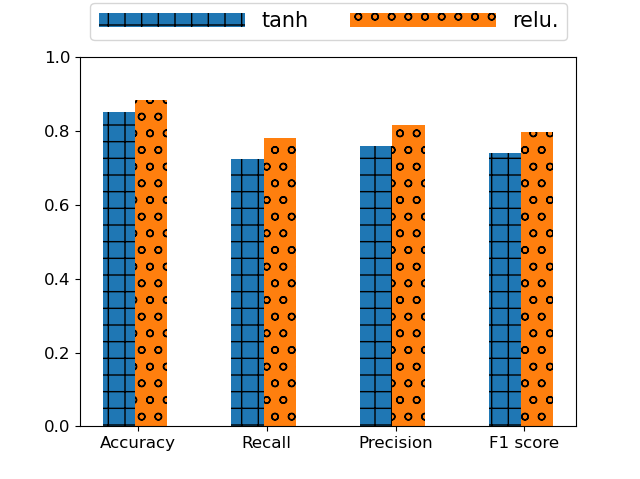}
    \caption{Effect of activation function}
    \label{fig:activation}
    \end{minipage}
\vspace{-0.2in}
\end{figure*}

\textbf{(Q4) Effect of Chosen Classifier.} Figure \ref{fig:algo} shows the effect of classifier choice on our architecture performance. The results clearly show that universal approximators neural networks significantly outperform the simple linear classifiers logistic regression and gradient boosting. This is also due to the fact that we have large number of features in our lower level learned feature representations. This answers \textbf{Q4}. 

\textbf{(Q5) Effect of Number of Neural Network Classifier Layers.} Figures \ref{fig:layers} and \ref{fig:activation} show the effect of number of layers and choice of activation function in the final neural network classifier on the model performance. As the results show, the number of layers do not have much effect on the performance but the activation function ``relu" outperforms ``tanh" due to the non-saturation of the calculated gradient thus accelerating the convergence of stochastic gradient descent (SGD). We also used simple SGD as the activation function, but the neural network did not converge. This answers \textbf{Q5}.

\section{Conclusion and Future Work}

We considered the challenging task of predicting DDIs from multiple sources. To this effect, we combined the data using embeddings created from images, SMILE strings from drug structures, and relationships between drugs. We presented an architecture that significantly outperforms strong baselines that learn from a single type of data.

More rigorous evaluation using larger data sets is an interesting direction. Potentially identifying novel DDIs is an exciting future research. Allowing for domain expert's knowledge could significantly boost the performance of the architecture and this can be achieved by considering the knowledge as constraints due to learning. Finally, understanding how it is possible to extract explanations of these interactions from the embeddings remains an interesting future direction.

\section{Acknowledgements}
%
%
%
\footnotesize
\bibliographystyle{splncs04}
\bibliography{test}
\end{document}